\pdfoutput=1
\documentclass[11pt]{article}
\usepackage{listings}
\usepackage[final]{acl}

\usepackage{times}
\usepackage{latexsym}
\usepackage{amsmath} 
\usepackage[T1]{fontenc}
\usepackage[utf8]{inputenc}

\usepackage{microtype}

\usepackage{inconsolata}

\usepackage{graphicx}
\usepackage{booktabs}

\title{AfriSpeech-MultiBench: A Verticalized Multidomain Multicountry Benchmark Suite for African Accented English ASR}

\author{
\begin{tabular}[t]{c}
\textbf{Gabrial Zencha Ashungafac\textsuperscript{1}}, 
\textbf{Mardhiyah Sanni\textsuperscript{1}}, 
\textbf{Busayo Awobade\textsuperscript{1}},\\
\textbf{Alex Gichamba\textsuperscript{1}}, 
\textbf{Tobi Olatunji\textsuperscript{1}}\\[3pt]
\textsuperscript{1} Intron Health \\[2pt]
\texttt{tobi@intron.io}
\end{tabular}
}
\usepackage[utf8]{inputenc}
\begin{document}
\maketitle
\begin{abstract}

Recent advances in speech-enabled AI, including Google's NotebookLM and OpenAI's speech-to-speech API, are driving widespread interest in voice interfaces globally. Despite this momentum, there exists no publicly available application-specific model evaluation that caters to Africa's linguistic diversity. We present \textbf{AfriSpeech-MultiBench}, the first domain-specific evaluation suite for over 100 African English accents across 10+ countries and seven application domains: Finance, Legal, Medical, General dialogue, Call Center, Named Entities and Hallucination Robustness. We benchmark a diverse range of open, closed, unimodal ASR and multimodal LLM-based speech recognition systems using both spontaneous and non-spontaneous speech conversation drawn from various open African accented English speech datasets. Our empirical analysis reveals systematic variation: open-source ASR models excels in spontaneous speech contexts but degrades on noisy, non-native dialogue; multimodal LLMs are more accent-robust yet struggle with domain-specific named entities; proprietary models deliver high accuracy on clean speech but vary significantly by country and domain. Models fine-tuned on African English achieve competitive accuracy with lower latency, a practical advantage for deployment, hallucinations still remain a big problem for most SOTA models. By releasing this comprehensive benchmark, we empower practitioners and researchers to select voice technologies suited to African use-cases, fostering inclusive voice applications for underserved communities.

%This document is a supplement to the general instructions for *ACL authors. It contains instructions for using the \LaTeX{} style files for ACL conferences.
%The document itself conforms to its own specifications, and is therefore an example of what your manuscript should look like. These instructions should be used both for papers submitted for review and for final versions of accepted papers.
\end{abstract}

\section{Introduction}

Automatic Speech Recognition (ASR) has become a foundational technology across numerous domains. In customer‑support environments, ASR powers real‑time call routing, intent detection, and agent assistance, substantially reducing response times and improving user satisfaction \cite{Wang2023CallCenter}. In healthcare, voice‑enabled digital scribes transcribe clinician-patient interactions on the fly, alleviating documentation burdens and cutting downstream transcription costs \cite{vanBuchem2021DigitalScribe}. Emerging applications in legal transcription \cite{DBLP:conf/icail/SaadanyBOW23}, financial trading desktops, and live subtitling further demonstrate the broad impact of ASR systems in both enterprise and consumer settings.

Selecting the optimal ASR model for a given task now often means choosing among powerful, pre‑trained \emph{foundation} systems rather than training bespoke models from scratch. Self-supervised models such as wav2vec 2.0  \cite{Baevski2020wav2vec} learn rich audio features from large amounts of unlabeled speech and can be applied in a zero‑shot or few‑shot manner, achieving near state‑of‑the‑art WER rates on standard benchmarks \cite{Baevski2020wav2vec}. Large multitask models such as Whisper \cite{pmlr-v202-radford23a}, trained on hundreds of thousands of hours of multilingual and multitask data, exhibit strong zero‑shot transfer across domains and languages without additional fine‑tuning \cite{pmlr-v202-radford23a}. However, computational budgets, latency requirements, and domain mismatches mean that one foundation model may outperform another depending on the target task, be it medical dictation, legal proceedings, or informal conversational speech.

Accented speech, particularly non-Western and underrepresented varieties, remains a persistent blind spot in mainstream evaluation suites. African accents, with their rich phonetic and prosodic diversity, often lead to significant word error rate disparities compared to North-American or British English \cite{Dossou2025AccentASR}. Without a dedicated benchmark, practitioners lack a reliable way to assess which off-the-shelf ASR system can meet accuracy, latency, or robustness requirements on African-accented speech.

Accordingly, we present a unified evaluation suite that benchmarks leading ASR systems, \textbf{AfriSpeech‑MultiBench}  in zero‑shot mode across medical, legal, conversational, entity-rich, call center, finance and robustness-diagnostic African-accented English speech (short, silence, and no-speech conditions). The suite supplies standardized test sets, and  transparent scoring protocols enabling practitioners to compare models and select the architecture most appropriate for their target application or for finetuning. We publicly release the benchmark suite on Hugging Face with CC BY-NC-SA 4.0  License \footnote{\url{https://huggingface.co/collections/intronhealth/afrispeechmultibench}}

\section{Related Work}

IrokoBench introduced a comprehensive text‑based evaluation across seventeen low‑resource African languages, revealing significant performance gaps between large language models and human competence on tasks such as natural language inference, reasoning and question answering \cite{adelani2025irokobench}.  The study underscores the necessity of domain‑specific evaluation: without targeted test suites, systematic deficiencies remain undetected.  

\par\medskip

Within automatic speech recognition (ASR), progress is often measured through the community‑maintained Open ASR Leaderboard, which continuously reports word‑error rate (WER) and real‑time factor on LibriSpeech \cite{panayotov2015librispeech}, TED‑LIUM 3 \cite{hernandez2018tedlium}, GigaSpeech \cite{chen2021gigaspeech}, VoxPopuli \cite{wang2021voxpopuli}, AMI \cite{carletta2005ami}, Earnings22 \cite{andrew2022earnings22}, SPGISpeech \cite{guo2022spgispeech}, and Common Voice \cite{ardila2020commonvoice}.  Although these datasets cover a range of domains, from read audiobooks to meeting‑room recordings, they remain dominated by North‑American and British English, providing limited insight into performance on African‑accented English.  

\par\medskip

Empirical investigations confirm the practical consequences of this imbalance. \citealp{koenecke2020racial} documented a twofold increase in WER for African American Vernacular English relative to Standard American English across multiple commercial recognizers. A global audit involving speakers from 171 birth countries observed the largest error rates for sub‑Saharan participants\cite{dichristofano2022accentgap}.  In the absence of African‑accented evaluation sets, leaderboard rankings therefore offer an incomplete picture for stakeholders on the continent.  

\par\medskip

Modern recognizers are architecturally diverse.  They include multilingual encoders such as Whisper \cite{pmlr-v202-radford23a} and XLS‑R, proprietary cloud services (Microsoft Azure Speech‑to‑Text, Google Speech‑to‑Text), Conformer‑based systems like Canary \cite{puvvada2024canary}
 and Parakeet \cite{rekesh2023fastconformer}, Speech‑Augmented Language Models (SALMs) \cite{chen2023salm}, and multimodal architectures such as SeamlessM4T \cite{schwenk2023seamlessm4t}.  Their heterogeneous training regimes and objectives complicate any attempt to infer accent robustness from results on existing benchmarks alone.  

\par\medskip

% we need to cite Edu-STT, NCHLT, and other papers
% use footnote for new intron datasets
% do not mention intron to protect anonymity

Several African‑accented corpora have been released to mitigate data scarcity. AfriSpeech‑200 provides roughly 200 hours of read speech from more than 100 indigenous accents \cite{olatunji2023afrispeech200}. AfriSpeech‑Dialog adds spontaneous two‑speaker conversations \cite{sanni2025afrispeechdialog}; AfriSpeech‑Parliament captures parliamentary debates \cite{intron2025parliament}; Med‑Convo‑Nig focuses on Nigerian clinical tele‑consultations \cite{intron2025medconvo}; Afri-Names targets named‑entity‑rich prompts \cite{intron2025afrinames}; and AfriSpeech‑Countries assembles cross‑regional accents under consistent recording conditions \cite{intron2025countries}.  Existing baseline evaluations do not cover modern speech recognition systems or lack broad application-specific results.

%published with these resources cover only a subset of the top‑performing models on the Open ASR Leaderboard, leaving open questions about comparative performance across domains and model families.  

\par\medskip

This study contributes three key advances.  First, 7 publicly available African‑accented corpora are harmonised into AfriSpeech-MultiBench, an evaluation suite spanning medical, legal, conversational , named‑entity‑rich and noise robustness speech.  Second, 18 contemporary recognizers covering multilingual, proprietary, Conformer‑based, SpeechLLMs and multimodal architectures are evaluated in zero‑shot mode, with  WER  reported.  Third, a fine‑grained error analysis disaggregates results by accent cluster, phonetic context and domain, elucidating systematic failure modes and informing future data collection and model selection.  

%\par\medskip
%The remainder of the paper is organised as follows.  Section \ref{sec:data} describes corpus harmonisation and evaluation split construction.  Section \ref{sec:models} outlines the models and inference settings.  Section \ref{sec:eval} details metrics and experimental protocol.  Section \ref{sec:results} presents quantitative findings, and Section \ref{sec:discussion} analyses error patterns and latency trade‑offs.  Section \ref{sec:conclusion} concludes with directions for accent‑aware ASR research.

\section{Benchmark Methodology}
\label{sec:data}

\subsection{Source Datasets}

We assemble seven corpora to form AfriSpeech‑MultiBench, covering diverse Anglophone African English accents. The distribution of sources is shown in Table \ref{tab:domain_distribution}.

%Existing ASR leaderboards focus on LibriSpeech, TED‑LIUM, GigaSpeech, VoxPopuli, AMI, Earnings22, SPGISpeech and Common Voice, yet none contain controlled samples of African‑accented English.  AfriSpeech‑MultiBench augments those benchmarks by unifying six publicly available corpora that together cover \textit{all regions of Africa}, five speech domains, and more than 108 distinct accents.  

\begin{table*}[ht!]
\centering
\small
\begin{tabular}{l l r r r r r}
\toprule
\textbf{Domain} & \textbf{Data Source} & \textbf{Samples} & \textbf{Hours} & \textbf{Countries} & \textbf{Accents} & \textbf{Speakers} \\
\midrule
Medical        &  Afri (clinical), Dialog (medical), Med.Conv  &  3651 & 29.88 & 10 &  95 & 519 \\
General        &  Afri (general), Dialog (general)  & 2741  & 13.06 & 9 &  84 &  455\\
Legal          &  Parl  &  8068   & 35.86  &  4 & --  & --  \\
Named Entities &    Names (names)      & 3121  &  2.18 & 3 & 6 & -- \\
Finance        &   Names (numbers), Names (commands)   &  3186   & 6.73  & 4 & 9 & -- \\
Call Center   &      Call (Private)   & 16  &  0.80 & 2  & 3 &  32 \\
Robustness   & Short Speech, No Speech, Intervening Silence       & 2067  &  3.74 & -  & - &  - \\

\midrule
\textbf{Total Unique} &                       & \textbf{20093} & \textbf{79.19} & \textbf{11} & \textbf{108} & \textbf{859} \\
\bottomrule
\end{tabular}
\caption{Domain-wise breakdown of the AfriSpeech-Multibench benchmark. Parentheses denote domain-specific subsets. Full names of the datasets - Afri:AfriSpeech, Dialog:AfriSpeech-Dialog, Med.Conv:Med-Conv-Nig, Names: AfriNames. The Call Center source is private and not disclosed.}
\label{tab:domain_distribution}
\end{table*}

\begin{itemize}
\setlength\itemsep{2pt}
\item \textbf{AfriSpeech-200}: (Afri) a ~200-hour, 67,577 clip dataset, 2,463 speakers across 120 indigenous accents from 13 African countries, spanning clinical and general domain read speech \cite{olatunji2023afrispeech200}.
\item \textbf{AfriSpeech‑Dialog:} (Diag) about 50 long-form medical and nonmedical conversational sessions with African-accented spontaneous English (about 7 hrs) \cite{sanni2025afrispeechdialog}.
\item \textbf{AfriSpeech-Parliamentary:} (Parl) A real-world noisy, multi-speaker dataset of transcribed parliamentary speech (about 35.86 hours, 8,068 clips) sampled from Nigeria, Ghana, South Africa, and Kenya. %Each sample is a 16-second audio segment featuring overlapping speakers and ambient noise—ideal for evaluating robustness in realistic conversational speech
\cite{intron2025parliament}.
\item \textbf{Med-Conv-Nig:} (Med.Conv) about 25 long-form simulated doctor-patient conversations capturing multispecialty clinical interactions in Nigeria, featuring both male and female speakers and rich in medical vocabulary  tailored for evaluating domain-specific ASR in healthcare settings 
%(4.2h) – real Nigerian tele‑consultations
\cite{intron2025medconvo}.
\item \textbf{AfriNames:} (Names) A read-speech corpus with subsets focused on African names (Name), numbers (Nums), and voice commands (Commands), e.g. "transfer \$500 to my HSBC account"; comprising 6,307 single‑speaker samples (about 8.92 hours), enriched with named entities and number utterances, spanning 12 distinct accents across four countries, particularly suited for evaluating ASR performance on entity-rich transcription tasks \cite{intron2025afrinames}
%(8.9h) – digit strings and named‑entity‑rich sentences for ASR–NER evaluation.
\item \textbf{AfriSpeech-Countries:} A mixture of AfriSpeech-200, AfriSpeech-Parliamentary, AfriNames and North African accented speech samples (Ctry-NA), totaling approximately 67 hours and 21,581 clips. The dataset spans seven African regions and includes both read and conversational speech. All samples are annotated by domain and country. 
\item \textbf{Afro-Call-Centers:} (Call) A private unreleased dataset capturing real-world agent-customer voice interactions rich in domain‑specific vocabulary across finance, health, and customer support domains
%(12.5h) – conversational clips from North Africa, complementing the seven‑region AfriSpeech‑Countries set.
\end{itemize}

% \begin{table}[t]
% \centering
% \small
% \begin{tabular}{l|r r r}
% \toprule
% \textbf{Dataset} & \textbf{Hrs } & \textbf{Speakers } & \textbf{Accents } \\ 
% \midrule
% AfriSpeech      & 18.68 & 750 & 108 \\ 
% Afri‑Diag       &   7.00 &   98 &  12 \\ 
% Parl            &  35.86 &   -- &   4 \\ 
% Med.Conv        &   4.20 &    11 &   1 \\ 
% Names           &   8.91 &    -- &  12 \\ 
% Countries (NA)*   & 4.61  &  --  &   7 \\ 
% \textbf{Total Unique}  & \textbf{79.26}& \textbf{859} & \textbf{108} \\ 
% \bottomrule
% \end{tabular}
% \caption{Corpus statistics (Test). Countries (NA) represents speech samples from Northern African  countries not included in other test sets which already have other African countries. Dashes represent statistics not provided in the original release of the datasets.}
% \label{tab:data_stats}
% \end{table}

\subsection{Domains Studied}
We define seven domain categories for evaluation with dataset details described in Table \ref{tab:domain_distribution}: 
\begin{itemize}
    \item \textbf{Medical:} health-related medical speech and clinician–patient dialogues.
    \item \textbf{General:} read-speech sourced from Wikipedia and non-spontaneous multispeaker dialogues.
    \item \textbf{Legal:} noisy parliamentary proceeding with overlapping speech.
    \item \textbf{Finance:} read speech enriched with numbers such as currencies, decimals, dates, measurements, locations, trading volumes, and financial institutions.
    \item \textbf{Call Center / Customer Support:} real-world agent–customer interactions
    \item \textbf{Named-Entities:} Named‑Entity‑Rich General clips with dense mentions of African person names, locations, organizations, and dates
    \item \textbf{Noise Robustness:} This diagnostic subset evaluates ASR stability under challenging acoustic conditions, including short utterances (under 3.5\,s) from AMI, VoxPopuli, AfriSpeech, and AfriNames datasets. It also includes \textit{Intervening Silence} clips from AfriNames with deliberate pauses to test contextual continuity, and a \textit{No Speech} subset from AfriSpeech-Parliament to measure false-trigger resistance when no speech is present.

\end{itemize}

\subsection{Models}
\label{sec:models}
We evaluate 19 modern ASR systems partly sourced from the top twenty entries on the Hugging Face Open ASR Leaderboard (snapshot: July 2025)\footnote{Leaderboard
URL: \url{https://huggingface.co/spaces/hf-audio/open_asr_leaderboard}.} categorized into model families representing architectural breadth Conformer, RNN‑T, CTC, transducer hybrids, and speech‑augmented language models (SpeechLLMs) and include both fully open‑source
checkpoints and proprietary services already deployed in commercial workflows.

\begin{table}[ht]
\centering
\small
\begin{tabular}{p{1.5cm} l l l}
\toprule
\textbf{Architecture} & \textbf{Model} & \textbf{Size} \\
\midrule
Conformer & Nvidia Parakeet‑tdt‑0.6B‑v2   & 0.6B \\
 & Nvidia Parakeet‑tdt‑1.1B      & 1.1B  \\
   & Nvidia Parakeet‑rnnt‑1.1B     & 1.1B \\
   & Nvidia Canary‑1B‑flash        & 1B   \\
   \hline
Whisper  
& OpenAI Whisper‑large‑v3   & 1.54B  \\
 & Distil‑Whisper‑v3.5    & 756M  \\
 & Nyra Health CrisperWhisper    & 1.54B   \\
 \hline
SpeechLLMs & IBM Granite‑3.3‑2B     & 2B   \\
 & Mistral Voxtral-Mini-3B      & 3B   \\
 & Nvidia Canary‑Qwen‑2.5B       & 2.5B  \\
 & Microsoft Phi‑4 MM‑Instruct  & 5.6B   \\
 \hline
Proprietary & Intron‑Sahara     & --   \\
 & Intron‑Sahara-V2     & --   \\
 & OpenAI GPT‑4o Transcribe      & -- \\
 & Google Gemini‑2.0 Flash & --  \\
 & Google Gemini‑2.5 Flash & --  \\
 & AWS Transcribe         & --   \\
 & Microsoft Azure Speech & -- \\
 & Google Chirp v3 & -- \\
\bottomrule
\end{tabular}
\caption{Descriptions of evaluated models, including model size, core architecture, and provider. Model sizes are in billions (B) of parameters when known.}
\label{tab:model_descriptions}
\end{table}

\begin{itemize}
\item \textbf{NVIDIA's open models:} Open‑source ASR models based on the FastConformer \cite{rekesh2023fastconformer} such as the Parakeet variants: CTC, RNN-T and TDT \cite{galvez24_interspeech} in sizes of 0.6B and 1.1B, and the 1 billion parameter Canary-flash model pairing a FastConformer encoder with a transformer decoder \cite{puvvada2024canary}.
\item \textbf{Whisper Variants:} Transformer encoder decoder models based on Whisper \cite{pmlr-v202-radford23a}. We consider the variants: Whisper-large-v3 \cite{pmlr-v202-radford23a}, Distil-Whisper-v3.5\footnote{\url{https://huggingface.co/distil-whisper/distil-large-v3.5}}, and CrisperWhisper \cite{zusag24_interspeech}.
\item \textbf{Open SpeechLLMs:} Multimodal LLMs and Speech-Augmented LLMs including IBM Granite-3.3-2B\footnote{\url{https://huggingface.co/ibm-granite/granite-speech-3.3-2b}}, Phi-4 Multimodal Instruct \cite{abdin2024phi4technicalreport}, Nvidia Canary‑Qwen\footnote{\url{https://huggingface.co/nvidia/canary-qwen-2.5b}}, and Mistral's Voxtral Mini-3B \cite{liu2025voxtral}.
\item \textbf{Proprietary cloud ASR services:} OpenAI's GPT-4o transcribe\footnote{\url{https://platform.openai.com/docs/models/gpt-4o-transcribe}}, Google's Gemini-2.0- \& Gemini-2.0- flash\footnote{\url{https://cloud.google.com/vertex-ai/generative-ai/docs/models/gemini/2-0-flash}}, Google's Chirp V3 \footnote{\url{https://cloud.google.com/speech-to-text/v2/docs/chirp_3-model}}, AWS Transcribe\footnote{\url{https://aws.amazon.com/transcribe/}}, Azure Speech Recognition\footnote{\url{https://azure.microsoft.com/en-us/products/ai-services/ai-speech}} and Intron Sahara (V1 and V2)\footnote{\url{https://www.intron.io/}}. Models are evaluated in zero‑shot mode, with neither demonstrations \cite{min-etal-2022-rethinking} nor domain-specific fine‑tuning.
%\item \textbf{Intron:} 
\end{itemize}

This broad selection of modern ASR systems facilitate an empirical comparison between commercially deployed services and publicly available checkpoints, capturing the architectural and commercial diversity of leading ASR systems, providing a realistic basis for accent‑aware model selection.

\subsection{Evaluation Protocol}
\label{sec:eval}

\begin{itemize}
    \item Primary metric: Word Error Rate (WER) measured per model, per domain, per country, and per dataset.
    \item Error analysis: Breakdown by domain, accent group (native vs non‑native), named‑entity errors, noise robustness; Open‑source vs proprietary models, unimodal vs multimodal, large vs compact variants.
    %\item Cross‑model comparisons: Open‑source vs proprietary, Conformer vs Whisper vs multimodal, large vs compact variants.
    % \item %Reporting: Tabular and visualization (WER vs latency), disaggregated trends to highlight domains or accents where model families perform best or degrade.
    % \item Latency: Real‑Time Factor (RTF) or inference time normalized for model size and hardware.
\end{itemize}

%This section details the signal and transcript processing steps applied before scoring, together with the metric used to compare the fourteen systems from Section~\ref{sec:models} on the test sets described in Section~\ref{sec:data}.

%Table~\ref{tab:wer_table} organises the empirical outputs of our benchmarking pipeline.  Each column corresponds to one of the widely used leaderboard test sets (LibriSpeech, TED‑LIUM 3, GigaSpeech, \ldots) or to a domain‑specific subset of AfriSpeech‑MultiBench.  Each row lists one of the fourteen systems introduced in Section \ref{sec:models}.  The table therefore allows readers to compare

%\begin{comment}
%    \begin{itemize}\setlength\itemsep{2pt}
%\item \textit{cross‑benchmark accuracy}, by %following a model across columns
%      rooted in different recording conditions and domains;
%\item \textit{architectural variation}, by scanning down a single benchmark
 %     column—from Conformer pipelines to Transformer–decoder and SALM models;
%\item \textit{production readiness}, via the inclusion of proprietary cloud
%      APIs alongside fully open‑source checkpoints.
%\end{itemize}

%All scores are word‑error rates (WER; lower is better) computed with the normalisation and evaluation protocol detailed in Section \ref{sec:eval}.   A more nuanced interpretation of these results—covering domain sensitivity,  and accent robustness appears later in Section \ref{sec:discussion}.
%\end{comment}

\begin{table*}[ht]
\centering
\tiny
\begin{tabular}{l r r r r r r r r r r r r r r}
\toprule
\textbf{Model} & 
\multicolumn{7}{c}{\textbf{Open ASR Benchmarks}} & 
\multicolumn{7}{c}{\textbf{AfriSpeech-MultiBench}} \\
\cmidrule(lr){2-8} \cmidrule(lr){9-15}
& Lib-S & TED-3 & Giga & VoxP & AMI & Earn22 & SPGI 
& Afri & Diag & Parl & MedC & Names & Call & Rob \\
\midrule
Parakeet-tdt-0.6B-v2   & 1.69 & 3.38 & 9.74 & 5.95  & 11.16  & 11.15 & 2.17 & 30.20 & \textbf{11.23} & 18.45 & 29.41 & 41.88 & 20.96 & 40.89 \\
Parakeet-tdt-1.1B      & \textbf{1.40} & 3.59 & 9.52 & 5.49  & 15.87  & 14.49 & 3.16 & 28.45 & 15.14 & 27.14 & 29.98 & 45.66 & 25.26 & 44.62 \\
Parakeet-rnnt-1.1B     & 1.45 & 3.83 & 9.89 & \textbf{5.44}  & 17.01  & 13.94 & 2.93 & 28.18 & 15.08 & 26.75 & 30.59 & 46.70 & 28.93 & 90.30 \\
Canary-1B-flash        & 1.48 & 3.12 & 9.85 & 5.63  & 13.11  & 12.77 & 1.95 & 29.77 & 48.50 & 19.13 & 93.62 & 44.10 & 88.71 & 51.32 \\
Whisper-large-v3       & 2.01 & 3.86 & 10.02 & 9.54 & 15.95  & 11.29 & 2.94 & 26.49 & 13.49 & 19.99 & 31.76 & 43.23 & 24.69 & 33.79 \\
Distil-Whisper-v3.5    & 2.37 & 3.64 & 9.84 & 8.04  & 14.63  & 11.29 & 2.87 & 27.58  & 11.50 & 18.00 & 30.41 & 45.80 & 21.65 & 34.00 \\
CrisperWhisper         & 1.82 & 3.20 & 10.24 & 9.82 & \textbf{8.71} & 12.89 & 2.70 & 63.80 & 72.72 & 79.35 & 83.12 & 70.14 & 35.52 & 38.82 \\
IBM Granite-3.3-2B     & 1.64 & 4.12 & 11.05 & 6.55 & 10.22 & 13.86 & 3.96 & 34.38 & 99.59 & 20.67 & 96.30 & 49.51 & 27.10 & 45.86 \\
Voxtral (Mistral)      & 1.86 & -- & 10.04 & 6.78 & -- & 12.18 & 2.04 & 20.17 & 68.42 & 21.10 & 78.73 & 49.36 & 29.20 & 43.51 \\
Canary-Qwen-2.5B       & 1.61 & \textbf{1.90} & \textbf{9.43} & 5.66 & 10.19 & \textbf{10.45} & \textbf{1.90} & 29.87 & 96.64 & 18.18 & 97.89 & 42.91 & 39.09 & 41.15 \\
Phi-4 MM-Instruct      & 1.68 & 2.89 & 9.77 & 5.93 & 11.45 & 10.50 & 3.11 & 26.48 & 88.91 & 36.73 & 130.17 & 44.28 & 24.99 & 122.45 \\
Intron-Sahara       & -- & -- & -- & -- & -- & -- & -- & 16.35 & 14.26 & 15.41 & 27.92 & \textbf{8.17}  & 20.08 & 18.91 \\
Intron-Sahara V2       & -- & -- & -- & -- & -- & -- & -- & \textbf{11.83} & 12.02 & \textbf{13.01} & \textbf{18.09} & 11.66 & \textbf{13.45} & \textbf{7.86} \\
GPT-4o Transcribe      & -- & -- & -- & -- & -- & -- & -- & 24.66 & 15.03 & 64.39 & 30.80 & 52.49 & 23.20 & 33.59 \\
Google Gemini-2.0 Flash& -- & -- & -- & -- & -- & -- & -- & 27.80 & 12.02 & 20.51 & 27.59 & 50.12 & 22.39 & 33.91 \\
Google Gemini-2.5 Flash& -- & -- & -- & -- & -- & -- & -- & 26.54 & 12.48 & 19.42 & 27.3 & 37.81 & 23.41 & 32.04 \\
AWS Transcribe         & -- & -- & -- & -- & -- & -- & -- & 32.77 & 14.02 & 18.50 & 30.08 & 36.70 & 23.51 & 33.98 \\
Azure Speech Recognition & -- & -- & -- & -- & -- & -- & -- & 28.41 & 13.29 & 18.75 & 29.17 & 35.69 & 24.95 & 32.61 \\
Google Chirp V3        & -- & -- & -- & -- & -- & -- & -- & 35.03 & 17.53 & 28.18 & 38.57 & 52.45 & 29.60 & 31.70 \\
\bottomrule
\end{tabular}

\caption{Word Error Rate (WER \%) for each model on standard open ASR benchmarks and subsets of the AfriSpeech‑MultiBench dataset. Dashes represent results that were not available. Full names of datasets: Lib‑S: LibriSpeech; TED‑3: TED-LIUM 3; Giga: GigaSpeech; VoxP: VoxPopuli; AMI: AMI Meeting Corpus; Earn22: Earnings22; SPGI: SPGISpeech; Afri: AfriSpeech-200; Diag: AfriSpeech-Dialogue; Parl: AfriSpeech-Parliamentary; MedC: Med-Conv-Nig; Nam: AfriNames; Call: Afro-Call-Centers.; Robustness a combination of No Speech from Afrispeech-Parliamentary;  Short Samples from AMI, VoxP and Afrispeech-Names; and Intervening Silence samples from AMI, VoxP, Afrispeech-Names}
\label{tab:overall-results}
\label{tab:wer_table}
\end{table*}

\begin{table}[ht]
\centering
\tiny
\begin{tabular}{l r r r r}
\toprule
\textbf{Model} & 
\textbf{Afri-Med} & 
\textbf{Diag} & 
\textbf{Med.Conv} & 
\textbf{Average} \\
\midrule
Parakeet-tdt-0.6B-v2      & 34.55 & \textbf{11.23} & 29.41 & 25.06 \\
Parakeet-tdt-1.1B         & 33.79 & 15.14 & 29.98 & 29.98 \\
Parakeet-rnnt-1.1B        & 33.45 & 15.08 & 30.59 & 30.59 \\
Canary-1B-flash           & 34.77 & 72.23 & 78.92 & 78.92 \\
Whisper-large-v3          & 32.59 & 17.22 & 31.76 & 27.19 \\
Distil-Whisper-v3.5       & 32.18 & 16.77 & 30.63 & 26.53 \\
CrisperWhisper            & 66.66 & 78.92 & 83.12 & 76.23 \\
IBM Granite-3.3-2B        & 40.28 & 99.53 & 96.30 & 78.70 \\
Voxtral (Mistral)         & 30.75 & 56.32 & 78.73 & 55.27 \\
Canary-Qwen-2.5B          & 32.04 & 93.08 & 97.92 & 74.35 \\
Phi-4 MM-Instruct         & 31.74 & 88.91 & 130.17 & 83.61 \\
Intron-Sahara             & 19.20 & 13.44 & 29.10 & 19.46 \\
Intron-Sahara V2          & \textbf{15.24} & 12.02 & \textbf{18.51} & \textbf{15.25} \\
GPT-4o Transcribe         & 28.54 & 15.03 & 30.80 & 24.79 \\
Google Gemini-2.0 Flash   & 31.13 & 12.02 & 27.59 & 23.58 \\
Google Gemini-2.5 Flash & 30.66 & 12.48 & 27.30 & 23.48 \\
AWS Transcribe            & 42.22 & 14.02 & 30.08 & 28.77 \\
Azure Speech Recognition  & 32.90 & 13.29 & 26.17 & 24.12 \\
Google Chirp V3           & 39.60 & 17.53 & 38.57 & 31.90 \\
\midrule
\textbf{Average}          & 34.59 & 39.51 & 53.83 & 42.89 \\
\bottomrule
\end{tabular}

\caption{Word Error Rate (WER \%) for each model on the medical domain subsets of AfriSpeech‑MultiBench, including clinical notes, medical dialogues, and doctor–patient conversations. Dataset full name mappings: Afri-Med: AfriSpeech Medical; Diag: AfriSpeech-Dialogue; Med.Conv: Med-Conv-Nig.}
\label{tab:medical_wer}
\end{table}

\begin{table}[ht]
\centering
\tiny
\begin{tabular}{lccc}
\toprule
\textbf{Model} & \textbf{Name} & \textbf{Commands} & \textbf{Nums} \\
\midrule
Parakeet-tdt-0.6B-v2      & 65.55 & 32.65 & 22.57 \\
Parakeet-tdt-1.1B         & 76.44 & 33.67 & 26.47 \\
Parakeet-rnnt-1.1B        & 75.78 & 35.36 & 26.66 \\
Canary-1B-flash           & 75.69 & 30.05 & 20.15 \\
Whisper-large-v3          & 73.10 & 31.58 & 18.11 \\
Distil-Whisper-v3.5       & 68.15 & 37.28 & 15.28 \\
CrisperWhisper            & 70.14 & 70.35 & 71.18 \\
IBM Granite-3.3-2B        & 78.97 & 49.03 & 23.45 \\
Voxtral (Mistral)         & 69.17 & 41.77 & 25.42 \\
Canary-Qwen-2.5B          & 69.79 & 31.44 & 20.15 \\
Phi-4 MM-Instruct         & 78.09 & 104.13 & 51.28 \\
Intron-Sahara             & 24.24 & \textbf{1.81} & 14.81 \\
Intron-Sahara V2          & \textbf{12.40} & 7.92 & \textbf{4.27} \\
GPT-4o Transcribe         & 67.43 & 46.67 & 17.45 \\
Google Gemini-2.0-Flash      & 74.12 & 40.77 & 18.23 \\ 
Google Gemini-2.5-Flash & 67.44 & 36.89 & 16.43 \\[2pt]
AWS Transcribe            & 60.07 & 27.60 & 20.21 \\
Azure Speech Recognition                      & 67.15 & 23.42 & 22.43 \\
Google Chirp V3           & 84.91 & 40.22 & 24.83 \\
\bottomrule
\end{tabular}

\caption{Word Error Rate (WER \%) for each model on African named entites and Financial domain subsets of AfriSpeech-MultiBench. Dashes represent results that were not available.}
\label{tab:names}
\end{table}

\begin{figure*} [hbt!]
    \centering
    \includegraphics[width=1\linewidth]{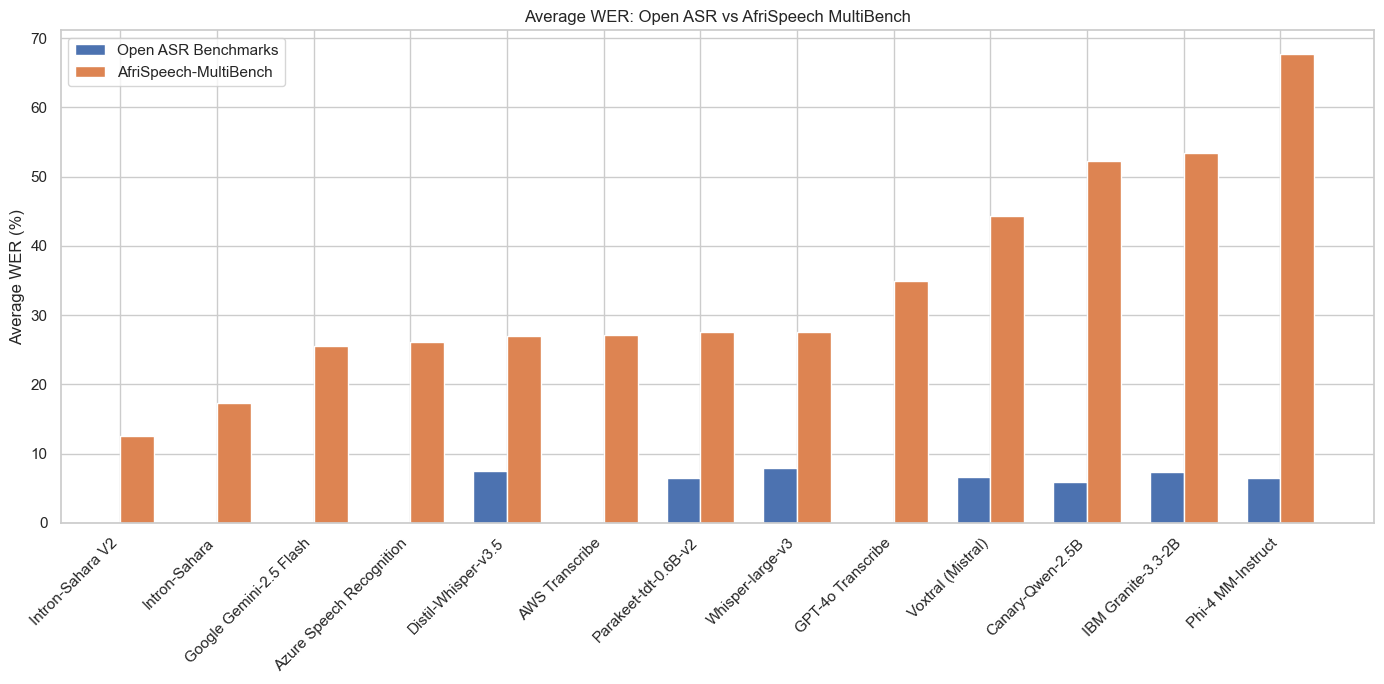}
    \caption{Average of Open ASR Leaderboard vs AfriSpeech-Multibench for top models from Table \ref{tab:overall-results}}
    \label{fig:overview}
\end{figure*}
\label{sec:discussion}

\begin{figure*}[hbt!]
    \centering
    \includegraphics[width=1\linewidth]{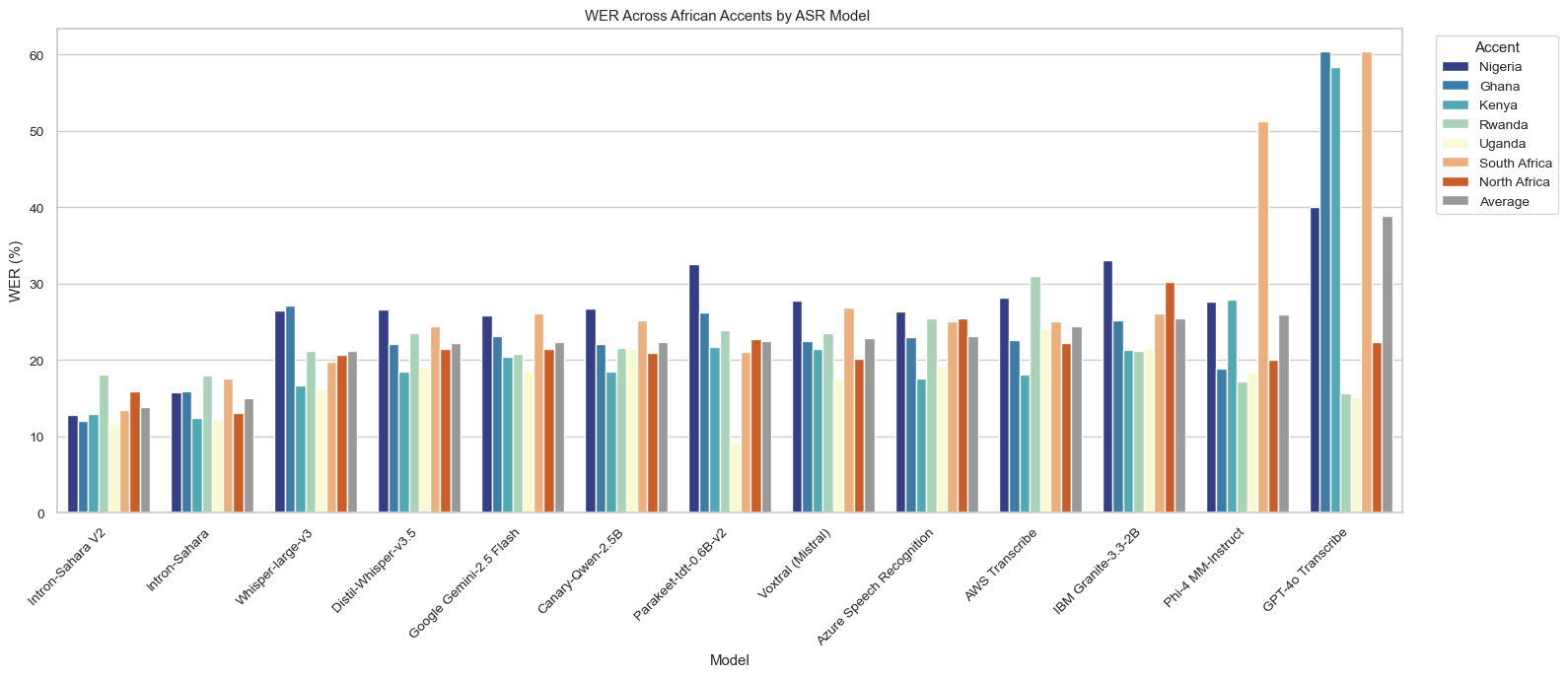}
    \caption{Word Error Rate (WER \%) for each model across different African English accents in AfriSpeech‑MultiBench. The average is computed across all listed accent categories.}
    \label{fig:enter-label}
\end{figure*}

\begin{table*}[ht]
\centering
\tiny
\begin{tabular}{l r r r r r r r r}
\toprule
\textbf{Model} & 
\textbf{Nigeria} & \textbf{Ghana} & \textbf{Kenya} & \textbf{Rwanda} & 
\textbf{Uganda} & \textbf{South Africa} & \textbf{North Africa} & \textbf{Average} \\
\midrule
Parakeet-tdt-0.6B-v2   & 32.60 & 26.27 & 21.78 & 23.92 &  \textbf{9.38} & 21.08 & 22.76 & 22.54 \\
Parakeet-tdt-1.1B      & 28.65 & 22.08 & 21.21 & 20.39 & 21.35 & 25.96 & 24.13 & 23.40 \\
Parakeet-rnnt-1.1B     & 32.76 & 23.69 & 23.19 & 24.71 & 23.08 & 27.59 & 24.42 & 25.63 \\
Canary-1B-flash        & 29.00 & 25.06 & 18.25 & 20.39 & 23.65 & 26.30 & 24.04 & 23.81 \\

Canary-Qwen-1B         & 28.10 & 23.90 & 19.20 & 20.60 & 20.00 & 24.80 & 22.10 & 22.81 \\

Whisper-large-v3       & 26.53 & 27.22 & 16.70 & 21.18 & 16.16 & 19.85 & 20.72 & 21.19 \\
Distil-Whisper-v3.5    & 26.69 & 22.16 & 18.51 & 23.53 & 19.22 & 24.45 & 21.43 & 22.28 \\
CrisperWhisper         & 74.50 & 80.99 & 74.72 & 40.00 & 58.29 & 72.11 & 62.64 & 66.18 \\
IBM Granite-3.3-2B     & 33.05 & 25.27 & 21.33 & 21.18 & 21.55 & 26.16 & 30.25 & 25.54 \\
Voxtral (Mistral)      & 27.84 & 22.49 & 21.50 & 23.53 & 17.57 & 26.96 & 20.17 & 22.87 \\
Canary-Qwen-2.5B       & 26.82 & 22.10 & 18.49 & 21.57 & 21.45 & 25.19 & 20.99 & 22.37 \\
Phi-4 MM-Instruct      & 27.73 & 18.86 & 27.92 & 17.25 & 18.49 & 51.26 & 20.03 & 25.93 \\
Intron-Sahara          & 15.85 & \textbf{15.93} & \textbf{12.48} & 18.04 & 12.26 & \textbf{17.65} & \textbf{13.14} & 15.05 \\
Intron-Sahara V2       & \textbf{12.83}  & 12.02 & 13.01 & 18.09 & 11.66 & 13.45 & 15.98 & \textbf{13.86} \\
GPT-4o Transcribe      & 40.03 & 60.41 & 58.38 & \textbf{15.69} & 15.22 & 60.43 & 22.40 & 38.94 \\
Google Gemini-2.0 Flash& 26.47 & 24.54 & 21.29 & 21.57 & 19.61 & 27.59 & 22.11 & 23.31 \\
Google Gemini-2.5 Flash& 25.90 & 23.10 & 20.40 & 20.85 & 18.70 & 26.10 & 21.50 & 22.36 \\

Google Chirp V3         & 27.10 & 24.00 & 21.20 & 21.90 & 19.60 & 27.30 & 22.60 & 23.67 \\

AWS Transcribe         & 28.16 & 22.59 & 18.18 & 30.98 & 24.11 & 25.05 & 22.23 & 24.47 \\
Azure Speech Recognition & 26.41 & 23.01 & 17.59 & 25.49 & 19.31 & 25.10 & 25.46 & 23.20 \\
\midrule
\textbf{Average}       & 30.99 & 27.54 & 24.36 & 23.01 & 21.25 & 29.65 & 24.38 & 25.31 \\
\bottomrule
\end{tabular}

\caption{Word Error Rate (WER \%) for each model across African accents in AfriSpeech-MultiBench, extended to include Voxtral (Mistral) and Intron-Sahara V2.}
\label{tab:countries_wer}
\end{table*}

\begin{figure*}[!t]
    \centering
    \includegraphics[width=\textwidth]{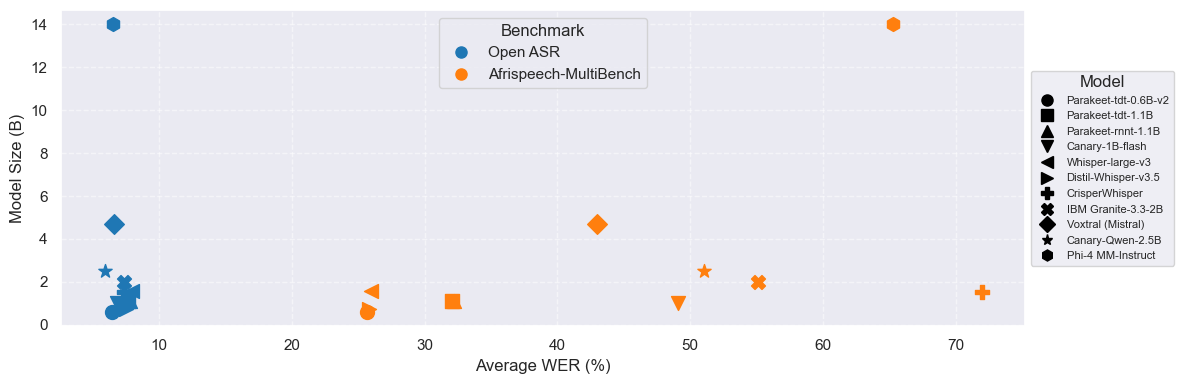}
    \caption{Model Sizes vs.\ Performance on Open ASR Benchmark (Blue) and AfriSpeech-Multibench (Orange)}
    \label{fig:open_asr_model_size_perf}
\end{figure*}

\section{Experiments}
\begin{itemize}
    \item Dataset splits: We use held‑out test sets per corpus, ensuring some accents appear only in testing to evaluate zero‑shot generalization (e.g. 41 accents exclusively in test partition of AfriSpeech‑200) 
    %\item Maximize Diversity: Ensure representation across domains, countries, and clean vs. noisy conversational speech.
    \item Transcript Pre- and Post-processing: Model-specific transcript pre- and post-processing (described in Appendix section \ref{transcript_processing}) normalized inputs, removed filler words, and mapped number words to their digit form, e.g. "twenty" to "20" and "first" to "1st". 
    \item Inference setup: Uniform audio input preprocessing (16 kHz mono, no diarization) with default hyperparameters and decoding settings for ASR models and proprietary API calls. Local runs were on single T4 GPU (16GB).
    \item Prompting:
    We use consistent prompts for open and closed LLMs, e.g. "Transcribe this ENGLISH audio". Prompt details are provided in Appendix section \ref{append_prompt}.  
\end{itemize}

We provide results for single runs.

\section{Results}
\label{sec:results}

\subsection{Overall Results}
As shown in Table~\ref{tab:overall-results}, model performance on widely used ASR benchmarks such as LibriSpeech, TED-LIUM, and AMI does not reliably translate to accuracy on African-accented or domain-specific speech. Models that dominate global leaderboards exhibit substantial degradation under accent, noise, and contextual variability characteristic of African datasets. For instance, leading open-source systems like \textbf{Parakeet-tdt-0.6B-v2} and \textbf{Whisper-large-v3}, which achieve sub-4\% WER on LibriSpeech, experience error rates between 30-45\% on general African speech and exceed 70\% on medical dialogue or named-entity-rich inputs within \textbf{AfriSpeech-MultiBench}. 

This discrepancy holds consistently across architecture families including transformer-based, transducer, and instruction-tuned multimodal models highlighting a persistent generalization gap of roughly 2-5$\times$ between leaderboard metrics and African deployment conditions. Notably, these errors compound in domain-specific contexts where pronunciation diversity, code-switching, and out-of-vocabulary proper nouns are prevalent.

In contrast, \textbf{Intron-Sahara V2} a successor to \textbf{Intron-Sahara} and a regionally tuned model trained on diverse African English data demonstrates markedly superior transferability. It achieves WERs below 15\% across all benchmarked domains, with a modest increase to 18.09\% in medical conversation, still outperforming all other models by a wide margin.

Beyond accuracy, \textbf{Intron-Sahara V2} exhibits exceptional robustness under challenging acoustic conditions. Across the robustness diagnostic subsets (\textit{Silence}, \textit{Short Samples}, and \textit{Intervening Silence}), it consistently delivers the lowest error and false-trigger rates, underscoring its resilience to pause-filled, truncated, or low-energy speech. These findings suggest that regionally tuned ASR systems can close much of the performance gap on African speech, outperforming larger yet globally trained models across both accuracy and robustness dimensions.
  %These findings highlight the limitations of global leaderboards and emphasize the importance of localized, domain-specific benchmarks like AfriSpeech-MultiBench for guiding real-world ASR deployment in Africa.

%Across all subsets of AfriSpeech‑MultiBench (AfriSpeech‑general, AfriSpeech‑dialogue, Parliamentary, Med Conversation, Named Entities, Non‑native country, Call Center), average WER ranges widely: Conformer‑based open models achieve ~20–30\% on scripted/clean data but degrade to ~40–60\% on dialogue and entity‑rich subsets. Proprietary services (OpenAI, Google Gemini, AWS, Azure) generally deliver lower WERs (~15–25\%) in dialogue and call center settings, though variability is observed across accents and domains.

\subsection{Domain Performance}

\subsubsection{Medical}
As shown in Table~\ref{tab:medical_wer}, the medical domain remains one of the most challenging settings, with average WERs exceeding 40\% for most open and proprietary systems. \textbf{Intron-Sahara~V2} achieves the best overall performance across all three medical datasets  on \textit{Afri-Med}, \textit{Afri-Diag}, and  \textit{Med.Conv} yielding an overall average of 15.26\%. Open models such as \textbf{Whisper-large-v3} and \textbf{Parakeet-tdt-0.6B-v2} perform moderately (25-27\%), while proprietary systems like \textbf{Gemini-2.0}, \textbf{GPT-4o}, and \textbf{Azure} range from 24–30\%. Large multimodal LLMs such as \textbf{Phi-4~MM-Instruct} and \textbf{IBM~Granite} exceed 80\% WER, underscoring that leaderboard success on clean English benchmarks does not generalize to accented or domain-specific medical speech. These results highlight the advantage of regionally tuned, domain-adapted ASR systems for low-resource healthcare applications in Africa.

%\subsubsection{General}
%Whisper‑large‑v3 has a competitive general WER (about 26.5\% on read/general; about 13.5\% on dialogue). Parakeet‑tdt‑0.6B achieves about 30.2\% on read speech and about 11.2\% on general dialogue. Canary‑1B‑flash also shows strong dialogue performance (about 19.1\%). Proprietary systems typically yield about 12–14\% WER on general dialogue.

\subsubsection{Finance}
The finance domain-represented by the \textit{Afri-Names} subsets for numerals and spoken commands shows strong gains for regionally tuned ASR. \textbf{Intron-Sahara~V2} achieves a WER of 8.20\%, compared to 35-50\% for most open-source and proprietary systems. Lightweight conformer models such as \textbf{Parakeet-tdt-1.1B} and \textbf{Whisper-large-v3} reach around 43-46\%, while \textbf{Azure} and \textbf{AWS} achieve mid-30\% accuracy. This domain highlights Sahara’s superior handling of short, context-free utterances and accent-driven variations in number pronunciation critical in the financial domain.

\subsubsection{Names}
Performance on African named entities remains a major differentiator. As shown in Table~\ref{tab:overall-results}, \textbf{Intron-Sahara~V2} again leads with 12.4\% WER, compared to 40-70\% for open-source models and over 50\% for proprietary systems. The failure modes of large general-purpose LLMs (e.g., hallucinating or anglicizing names) emphasize the need for phonetic grounding and localized lexicons. Despite limited scale, Sahara’s region-specific acoustic and language modeling yields more accurate entity recovery.

\subsubsection{Legal}
Table~\ref{tab:overall-results} summarizes performance on the \textit{Parliamentary} dataset, which features overlapping speakers and high background noise. Among open-source systems, \textbf{Distil-Whisper-v3.5} performs best with 11.5\% WER, showing that smaller distilled variants can generalize better to real conversational overlap. \textbf{Intron-Sahara~V2} follows closely at 12.01\%, outperforming all proprietary and large multimodal models, whose WERs remain above 20\%. This demonstrates that domain-adapted encoders trained on regional conversational data can surpass even the largest general-purpose models under high-noise conditions.

\subsubsection{Call Center}
In the call-center domain, \textbf{Intron-Sahara~V2} again leads, achieving 13.45\% WER, followed by \textbf{Whisper-large-v3} at 24.7\% and \textbf{Parakeet-tdt-0.6B-v2} at 20.96\%. Proprietary models such as \textbf{GPT-4o Transcribe} and \textbf{Gemini-2.0} record around 22-23\% WER, showing that Sahara’s tuning for multi-speaker interaction, turn-taking, and accent robustness provides a tangible advantage in noisy, dialogue-heavy environments.

\subsubsection{Noise Robustness}
The \textbf{Noise Robustness} evaluation combines the \textit{Silence}, \textit{Short}, and \textit{Intervening Silence} diagnostics to assess model stability under pauses and non-speech conditions. Most global ASR systems degrade sharply (20--70\% WER), often hallucinating speech. In contrast, \textbf{Intron-Sahara V2} achieves 0 \% false triggers on silence, 10.15\% on short clips, and 11.23\% under pauses, outperforming \textit{Whisper-large-v3} and \textit{GPT-4o}. Detailed results are provided in Appendix \ref{noise_rob}
 
%OpenAI GPT‑4o Transcribe and Whisper‑large‑v3 both achieve ~22–24\% WER, while Intron‑Sahara and Parakeet‑0.6B score ~20.9\% and ~20.96\% respectively. Azure and AWS Transcribe fall in the ~22–25\% range. Larger Conformer variants exhibit slightly higher error rates (~25‑29\% WER).

%\subsection{Conversational vs Read-Speech performance}
%Table \ref{tab:overall-results} shows that Conversational datasets like AMI and Earnings22 are more challenging that read speech datasets like Librispeech and VoxPopuli. This trend holds in MultiBench where performance on 

%Open-source ASR models (Parakeet/Canary) perform well in scripted general and legal contexts (e.g. parliament speech) but degrade sharply in conversational or medical dialogue contexts. In contrast, Intron‑Sahara, tuned for African English clinical data, excels on dialogue and Med Convo subsets, illustrating the value of domain‑specific fine‑tuning.

\subsection{Accent and country variations}
As shown in Table \ref{tab:countries_wer} and Figure \ref{fig:enter-label}, most models show pronounced degradation in Nigeria, South Africa, and Ghana (about 30\%), relative to East and North Africa (about 24\%). Most models perform comparably  except GPT-4o and CrisperWhisper with WERs above 60\% and Sahara models with WER less than 15\%.

\subsection{Model size vs performance}
Figure \ref{fig:open_asr_model_size_perf} and Table \ref{tab:overall-results} show that, in a handful of domains, larger Speech LLMs (Granite, Phi-4, Voxtral, Canary-Qwen) only marginally outperform smaller architectures like conformer and Whisper variants half their sizes. In conversational speech, they are worse overall. Figure \ref{fig:open_asr_model_size_perf} indicates overall worse performance for open models with increasing size.

\section{Discussion}
This study yields a number of key insights that illuminate performance gaps and opportunities for advancing ASR systems in African settings:

\subsection{Global benchmarks misrepresent African realities.} Leading models like Whisper and Parakeet achieve WERs below 10\% on LibriSpeech and GigaSpeech, yet degrade to over 20-40\% on African-accented data in AfriSpeech-MultiBench. This mismatch underscores the limits of current leaderboards in guiding ASR adoption across low-resource geographies.

\subsection{Accent diversity drives large performance variance.} While models performed well on Kenyan and Ugandan English (average WERs as low as 12-18\%), WERs doubled or tripled for West African and North African accents-exceeding 25\% for many systems. This highlights the phonetic and prosodic diversity across the continent and the inadequacy of accent-agnostic training.

\subsection{Conversational speech remains a major bottleneck.} Compared to read speech, performance worsened significantly on conversational corpora AfriSpeech-Dialog Medical, Medical Conversations (Med Convo), and Parliamentary speech. These mirror Western benchmarks, where models also struggle on AMI and Earnings22 relative to LibriSpeech or SPGISpeech. However, the drop-off in African conversational domains is more severe, revealing compound challenges likely due to accent, prosody, and domain shift.

\subsection{Named entities and structured commands still confound models.} Most models scored above 40\% WER on the Afri-Names dataset, numbers, and financial voice commands, often failing to distinguish culturally unique or phonetically similar terms. This raises usability concerns in domains requiring accurate name capture or transactional integrity.

\subsection{Model size and architecture don’t predict reliability.} Smaller models like Parakeet-tdt-0.6B and Distil-Whisper sometimes matched larger peers on global benchmarks but showed inconsistent gains on African test sets. By contrast, Sahara a regionally optimized model consistently delivered best-in-class results across medical, legal, and conversational tasks.

\subsection{Benchmarking must evolve beyond average-case accuracy.} AfriSpeech-MultiBench enables fine-grained, domain-aware evaluation that reflects real-world deployment conditions. It provides not only model ranking, but also insight into where and why systems fail offering practical guidance for building domain and region-specific ASR solutions in healthcare, law, finance, and public service delivery across Africa.

\section{Conclusion}
This study set out to address the gap between global ASR benchmarks and real-world performance on African-accented, domain-specific speech. Through AfriSpeech-MultiBench, we reveal that top-performing models on standard datasets like LibriSpeech and TED-3 achieving sub-5\% WER can exhibit 5-10X higher error rates on African speech, especially in medical, financial, and conversational domains. These disparities are consistent across open-source and proprietary systems, highlighting persistent geographic, linguistic, and domain biases in existing ASR development and evaluation pipelines.

Our findings underscore the need for regionally grounded benchmarks and models. Intron-Sahara, a model trained with African-specific data, consistently outperformed global leaders across domains and accents, particularly in name recognition, doctor patient dialogue, and financial commands. By benchmarking 19 models across 8 African countries and 7 key domains, AfriSpeech-MultiBench provides actionable insights for building inclusive ASR systems. This work lays the foundation for future research and deployment efforts in healthcare, legal transcription, customer service, and multilingual voice applications across the African continent.

\section*{Limitations}

While AfriSpeech-MultiBench offers a broad and diverse benchmark across African-accented English, several limitations warrant consideration. First, despite including over 10 countries and six domains, the benchmark does not yet cover all major linguistic regions in Africa or fully represent under-resourced countries with limited public data availability. Certain domains such as manufacturing, education, and public safety are not currently included, and even within included sectors like healthcare and finance, dataset sizes remain modest compared to global corpora, which may limit fine-grained error analysis and generalization of results.

Secondly, some datasets used are proxies rather than fully representative of their target verticals. For instance, parliamentary proceedings may not fully capture the legal domain’s complexity, such as courtroom vernacular, legalese, or multilingual code-switching common in legal aid and judicial settings. Similarly, due to privacy constraints, customer support datasets from private call centers were not included, limiting direct benchmarking for commercial deployments. These gaps highlight both the urgent need and the opportunity for continued investment in domain-specific and geographically expansive data collection to build more comprehensive benchmarks for inclusive speech technologies.

\
\label{sec:conclusion}
%\section{Bib\TeX{} Files}
%\label{sec:bibtex}

%\section*{Acknowledgments}

% Bibliography entries for the entire Anthology, followed by custom entries
%\bibliography{anthology,custom}
% Custom bibliography entries only
\bibliography{bibfile}
\newpage
\appendix

\section*{Appendix}
\subsection*{Noise Robustness}
\label{noise_rob}
\begin{table}[ht]
\centering
\scriptsize
\setlength{\tabcolsep}{3pt}
\begin{tabular}{l@{\hspace{3pt}}r@{\hspace{3pt}}r@{\hspace{3pt}}r@{\hspace{3pt}}r}
\toprule
\textbf{Model} & \textbf{SIL} & \textbf{Short} & \textbf{Int.SIL} & \textbf{Avg.} \\
\midrule
Parakeet-tdt-0.6B-v2   & 73.12 & 34.50 & 15.05 & 40.89 \\
Parakeet-tdt-1.1B      & 78.31 & 33.20 & 22.35 & 44.62 \\
Parakeet-rnnt-1.1B     & 211.62 & 35.41 & 23.86 & 90.30 \\

Canary-1B-flash        & 71.02 & 54.56 & 30.39 & 51.32 \\
Whisper-large-v3       & 41.90 & 36.58 & 22.89 & 33.79 \\
Distil-Whisper-v3.5    & 43.31 & 37.38 & 23.28 & 34.00 \\
CrisperWhisper         & 53.62 & 36.07 & 24.76 & 38.82 \\
IBM Granite-3.3-2B     & 43.90 & 60.75 & 32.92 & 45.86 \\
Voxtral (Mistral)      & 40.77 & 57.57 & 32.18 & 43.51 \\
Canary-Qwen-2.5B       & 37.63 & 54.39 & 31.43 & 41.15 \\
Phi-4 MM-Instruct      & 154.05 & 136.60 & 75.71 & 122.45 \\
Intron-Sahara          & 25.49 & 21.47 & 9.77 & 18.91 \\
\textbf{Intron-Sahara V2} & \textbf{0.00} & \textbf{15.98} & \textbf{7.59} & \textbf{7.86} \\
GPT-4o Transcribe      & 44.48 & 37.80 & 18.48 & 33.59 \\
Google Gemini-2.0 Flash& 45.45 & 37.80 & 18.48 & 33.91 \\

Google Gemini-2.5 Flash & 44.27 & 19.964 & 31.875 & 32.04 \\

AWS Transcribe         & 43.45 & 35.36 & 21.12 & 33.98 \\
Azure Speech Recog.    & 44.27 & 32.52 & 21.04 & 32.61 \\
Google Chirp V3        & 31.37 & 40.62 & 23.10 & 31.70 \\
\midrule
\textbf{Average}       & 58.86 & 45.62 & 26.18 & 43.55 \\
\bottomrule
\end{tabular}

\caption{Robustness evaluation across \texttt{SIL}, \texttt{Short}, and \texttt{Int.SIL} subsets. The final column shows their mean (\texttt{Avg.}).}
\label{tab:robustness_appendix}
\end{table}

\section*{Pre- and Post-Processing}

\subsection*{Audio pre‑processing}
\label{audio_preprocsiing}
Audio files are used exactly as distributed by the source datasets; no further
segmentation or concatenation is performed.  
A single exception concerns the NVIDIA NeMo checkpoints
(\textit{parakeet‑*}, \textit{canary‑1B}), which require 16kHz mono input.
When a file is multi‑channel or sampled above 16kHz, it is down‑mixed to mono
and re‑sampled with \texttt{sox} prior to inference.  
All other engines (Whisper variants, API endpoints) accept the original
wave‑forms without modification.

\subsection*{Transcript pre‑processing}
\label{transcript_processing}
Reference and hypothesis strings undergo a three‑stage normalisation pipeline,
implemented exactly as in the public evaluation script:

\begin{enumerate}\setlength\itemsep{1pt}
\item \texttt{clean\_text} \textemdash\; lower‑cases, trims whitespace, removes
   punctuation, deletes 32 variants of \emph{[inaudible]}, and removes
   frequent filler words (\emph{uh}, \emph{hmm},\emph{mmhmm},\ldots).
\item \texttt{text\_to\_numbers} \textemdash\; maps number words
   (\emph{“twenty’’} $\rightarrow$ 20) and ordinal words
   (\emph{“first’’} $\rightarrow$ 1st) to their digit form.
\item \texttt{EnglishTextNormalizer} \textemdash\; applies the Whisper
   normaliser for final case‑folding and whitespace cleanup.
\end{enumerate}

A sentinel token \texttt{abcxyz} replaces empty strings to avoid undefined
denominators in word‑error calculations.

\subsection*{Post‑processing for Nemo models}

NeMo/Parakeet outputs include automatically generated punctuation.  Before the
three‑stage normaliser, inverse text normalisation is applied to restore
standard spacing around commas and periods, ensuring a fair comparison with
punctuation‑free reference strings.

\subsection*{Metric}
\label{apped_metric}

Word‑error rate (WER) is computed with \textsc{JiWER}

\[
\text{WER}(r,h)=\frac{S+D+I}{|r|},
\]

where \(S\), \(D\) and \(I\) count substitutions, deletions and insertions
needed to transform hypothesis \(h\) into reference \(r\).  

\subsection*{Prompting for Speech Augmented Language Models}
\label{append_prompt}
Default prompts for open source speech augmented language models where used:
\begin{itemize}
    \item Canary-Qwen-2.5B : \begin{lstlisting}
"Transcribe the following: {model.audio\_locator\_tag}", "audio": ["speech.wav"]
\end{lstlisting}
\item Mixtral (Voxtral-Mini-3B-2507): We used its apply\_transcription\_request function which takes an audio file and wraps it with inbuilt prompts for speech transcription.
\item Google Gemini 2.0 Flash: The required prompt according to Google API documentation was used, prompt = """
Transcribe this ENGLISH audio.
"""
\item Phi-4 Multimodal Instruct: <|user|><|audio\_1|>Transcribe the audio to text<|end|><|assistant|>
\end{itemize}
\end{document}